\documentclass[10pt,twocolumn,letterpaper]{article}

\usepackage{cvpr}
\usepackage{times}
\usepackage{epsfig}
\usepackage{graphicx}
\usepackage{amsmath}
\usepackage{amssymb}
\usepackage[ruled]{algorithm2e}
\usepackage{bm}
\usepackage{multirow}
\usepackage{subfig}
\usepackage{float}
\usepackage{mathrsfs}
\usepackage{array}
\usepackage{esvect}
\usepackage{cite}

\usepackage[pagebackref=true,breaklinks=true,letterpaper=true,colorlinks,bookmarks=false]{hyperref}

 \cvprfinalcopy 

\ifcvprfinal\fi
\begin{document}

\title{Identity Preserving Generative Adversarial Network for Cross-Domain Person Re-identification}
\author{Jialun Liu, Wenhui Li, Hongbin Pei\\
College of Computer Science and Technology, Jilin University, China\\
}


\maketitle

\begin{abstract}
Person re-identification is to retrieval pedestrian images from no-overlap camera views detected by pedestrian detectors. Most existing person re-identification (re-ID) models often fail to generalize well from the source domain where the models are trained to a new target domain without labels, because of the bias between the source and target domain. This issue significantly limits the scalability and usability of the models in the real world. Providing a labeled source training set and an unlabeled target training set, the aim of this paper is to improve the generalization ability of re-ID models to the target domain. To this end, we propose an image generative network named identity preserving generative adversarial network (IPGAN). The proposed method has two excellent properties: 1) only a single model is employed to translate the labeled images from the source domain to the target camera domains in an unsupervised manner; 2) The identity information of images from the source domain is preserved before and after translation. Furthermore, we propose IBN-reID model for the person re-identification task. It has better generalization ability than baseline models, especially in the cases without any domain adaptation. The IBN-reID model is trained on the translated images by supervised methods. Experimental results on Market-1501 and DukeMTMC-reID show that the images generated by IPGAN are more suitable for cross-domain person re-identification. Very competitive re-ID accuracy is achieved by our method.
\end{abstract}

\begin{figure}[]
\setlength{\abovecaptionskip}{-0.1cm}
\setlength{\belowcaptionskip}{-0.2cm}
\begin{center}
\includegraphics[width=1.0\linewidth]{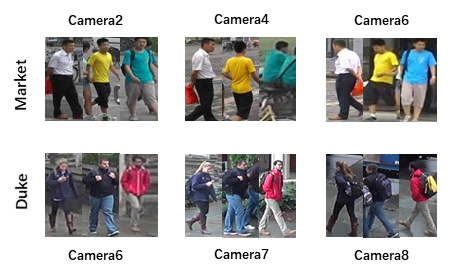}
\end{center}
\caption{Example images from 2,4,6 cameras of Market-1501 and 6,7,8 cameras of DukeMTMC-reID, respectively. The images with a same identity have different appearances in different camera views. }
\label{fig:carton}
\end{figure}

\section{Introduction}\label{Introduction}

As one of the most challenging problems in the field of computer vision, the re-ID task aims at searching for the relevant images from a gallery according to a query. The traditional person re-identification approaches \cite{koestinger2012large,liao2015person,zhang2016learning} are based on hand-crafted features. With the recent development of deep learning, many excellent deep learning-based supervised methods \cite{zheng2016person,sun2017svdnet,hermans2017defense,sun2017beyond,zhong2018camera} are proposed for the person re-identification task. Although those approaches has obtained a dramatic performance improvement, they have many limitations in the real-world mainly including 1)  the image labeling process is very expensive and impractical for supervised learning; 2) when re-ID models trained on a source domain and used on a target domain, the bias between the source and the target domain leads to performance degradation notably. Thus, it is critical to improve the generalization capacity of the supervised re-ID methods.
\par A common strategy to solve this problem is unsupervised domain adaptation \cite{hoffman2017cycada,long2015learning,tzeng2017adversarial,ghifary2014domain}, which assume that the source and the target domain contain a same set of classes. However, the assumption is not appropriate for the person re-ID task in which the source and the target domain have completely different persons (classes).

\par Very recently, few person re-ID methods\cite{deng2018image,wei2017person}  based on unsupervised domain adaptation are proposed, which use a generative adversarial network (GAN) to translate images from source domain to target domain. However, these methods only take into consideration the general gap between source domain and target domain but ignore the bias between source domain and target camera domains.  Actually, images captured by different cameras often have many obviously different styles. As Figure \ref{fig:carton} shows, the images with a same identity have different appearances in different camera views. Images captured by one camera can be regarded as a subdomain of target domain. Thereby, DukeMTMC-reID and Market1501 has 8 and 6 subdomains respectively because of their camera number. In the real world, the distribution of one subdomain may greatly differ from the distribution of the other ones, because the types of camera and scenes for image acquisition are different. In this case, it is improper to take the target domain as a whole. The better solution of domain adaptation is to reduce the \emph{bias} between source domain and each subdomain (camera domain) in target domain.

\par In this paper, we propose a novel and efficient unsupervised domain adaptation approach named Identity Preserving Generative Adversarial Network (IPGAN), which can generate images for target camera domains learning. The design of IPGAN is motivated by three aspects. Firstly, A significant motivation is to reduce the \emph{bias} between source and each target camera domains. Secondly, although the style of images in source domain is changed, the translated image should preserve the same identity information with its corresponding original image. Thirdly, the computational cost of cross domain person re-ID should not be very expensive because the dataset usually very large. To achieve the above three objectives, we design IPGAN which is inspired by StarGAN \cite{choi2017stargan}. In IPGAN, we design a novel semantic discriminator to implement the semantic constraint that the identity information of images from source domain keeps invariance before and after translation. Such a semantic discriminator brings more challenges to a generator in GAN framework. Using IPGAN, we can create a new dataset in an unsupervised manner, which inherits the labels from  source domain and has the style of target camera domains.
Then, we train the reID model on the new dataset in a supervised way.

\par For the person re-ID task, we further propose IBN-reID model inspired by \cite{pan2018two}.
The model can eliminate appearance variance in shallow layers, and holds discrimination of the learned features in deep layers. In the model, instance normalization and batch normalization are integrated. Instance normalization provides visual and appearance invariance, while batch normalization accelerates training and preserves discriminative features. The IBN-reID model has better generalization ability than the baseline model \cite{deng2018image}, especially in the case of deploying the model trained on source domain to target domain without any domain adaptation.

In summary, this paper has the following contributions:
\begin{itemize}
\item To solve domain adaptation in person re-ID task, we propose IPGAN, a novel and efficient unsupervised learning methods. It works by mapping the images from source domain to target camera domains with one single model meanwhile, preserving the identity information of images from source domain.
\item we present IBN-reID model which intuitively induces appearance invariance into re-ID model. It has better generalization ability than baseline model.
\item Many experimental results show that the proposed methods achieve very competitive re-ID accuracy and it is efficient and applied.
\end{itemize}

\begin{figure*}[t]
\setlength{\abovecaptionskip}{-0.2cm}
\setlength{\belowcaptionskip}{-0.3cm}
\begin{center}
\includegraphics[width=1\linewidth]{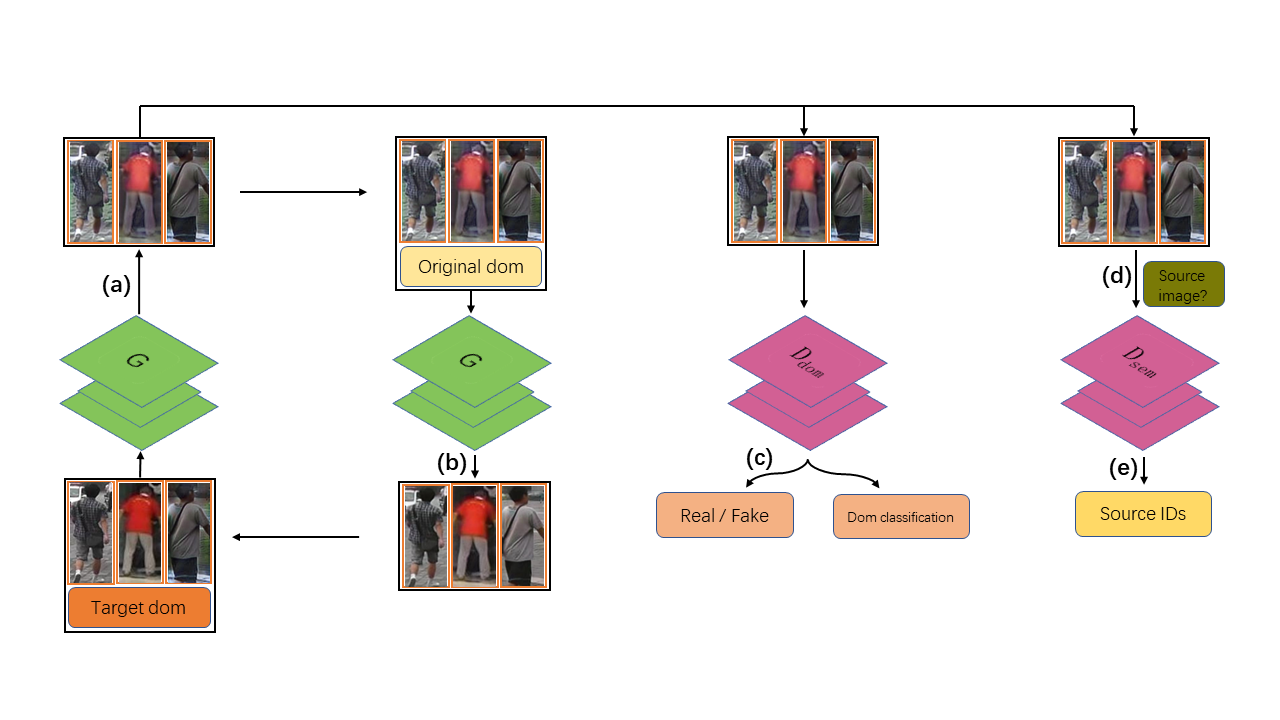}
\end{center}
\caption{Overview of IPGAN, consisting of three modules, a generator G, a domain discriminator $D_{dom}$, and a semantic discriminator $D_{sem}$. (a) the generator G learns to focus on the domain label (target cam label) to perform image-to-image translation. (b) G attempts to reconstruct images with original domain label and fake image. (c) the domain discriminator $D_{dom}$ learns to distinguish between real and fake images and minimize the classification error for the known label. (d) if the fake images is generated based on images from source domain images, input it in the semantic discriminator $D_{sem}$. (e) $D_{sem}$ claculates identity semantic loss.}
\label{fig:IPGAN-framework}
\end{figure*}

\section{Related Work} \label{Related Works}
\textbf{Generative Adversarial Networks.}
Generative adversarial networks (GANs) \cite{Goodfellow2014Generative} has shown remarkable performance improvement in various computer vision tasks, especially image-to-image translation, in recent years. For image-to-image translation task, pix2pix \cite{isola2017image} uses a conditional GANs to learn mappings from input to output images by combining adversarial loss and $L_{1}$ loss. However, this method needs paired data to train its model. For unpaired image-to-image translation, several methods are proposed \cite{zhu2017unpaired,kim2017learning,yi2017dualgan,zheng2017unlabeled}. UNIT \cite{liu2017unsupervised} combines variational autoencoders \cite{kingma2013auto} and CoGAN \cite{liu2016coupled}, in which the two generators share same weights. CycleGAN \cite{zhu2017unpaired} and DiscoGAN \cite{kim2017learning} use a cycle consistency to preserve key attributes. However, all the aforementioned frameworks only consider the mapping from source domain to target domain. Different form them, we propose a new framework which can translate images from source domain to each target camera domains using only a single model and be used to improve the performance of cross domain person re-ID.

\textbf{Unsupervised domain adaptation.}
Our work is one of unsupervised domain adaptation method where the labels of target images are unavailable \cite{long2015learning,Chen2018Person,Ajakan2014Domain}. In existing methods, a popular idea is  to reduce the divergence between source domain and target domain \cite{sun2016return,gretton2012kernel,Ganin2015Unsupervised}. CORAL \cite{sun2016return} gets good performance in various computer vision tasks by aligning the mean and covariance of two data distributions of source and target domain. By introducing the Maximum Mean Discrepancy (MMD), \cite{gretton2012kernel} try to reduce the MMD distance between the two domains. DANN \cite{Ganin2015Unsupervised} integrates the gradient reversal layer (GRL) into the standard architecture to ensure the similar distribution of features on both domains. There are many methods which attempt to produce fake-labels for the unlabeled samples \cite{Rohrbach2013Transfer,Saito2017Asymmetric,sener2016learning,zhu2006semi}. For instance, \cite{zhu2006semi} trains a classifier on labeled and fake-labeled samples to the predict labels of unlabeled samples. In \cite{Saito2017Asymmetric}, three classifiers are modelled to generate fake-labels for samples in target domain. Recently, many domain adaptation approaches \cite{hoffman2017cycada,bousmalis2017unsupervised} based on Generative Adversarial Networks \cite{Goodfellow2014Generative} are proposed. CyCADA \cite{hoffman2017cycada} achieves domain adaptation at both the pixel-level and feature-level by utilizing pixel cycle consistency and semantic losses. However, the above domain adaption approaches can not be used for the cross domain person re-ID task, because they assume that the source and target domain have same class labels. Actually, in the community of person re-ID, identities(classes) of different datasets have no overlap.

\textbf{Unsupervised person re-ID.}
 Most existing re-ID models are based on supervised learning \cite{koestinger2012large,Xiong2014Person,Wang2014Person,zhang2016learning,Wang2016Joint,Xiao2016Learning,Chen2016A,subramaniam2016deep}. These models suffer from poor scalability in the real-world environment. To solve this scalability issue and improve the generalization ability, unsupervised methods based on hand-crafted features \cite{Farenzena2010Person,Kodirov2016Person,Rui2017Person,Gray2008Viewpoint,liao2015person,cheng2011custom,kodirov2015dictionary,Wang2014Person,bazzani2013symmetry} can be chosen and applied. These methods aim to design or learn robust feature for person re-ID. But, they ignore the distribution of samples in the dataset and yield much weaker performance on large-scale dataset than supervised learning methods.

\begin{figure*}[t]
\setlength{\abovecaptionskip}{-0.2cm}
\setlength{\belowcaptionskip}{-0.3cm}
\begin{center}
\includegraphics[width=1\linewidth]{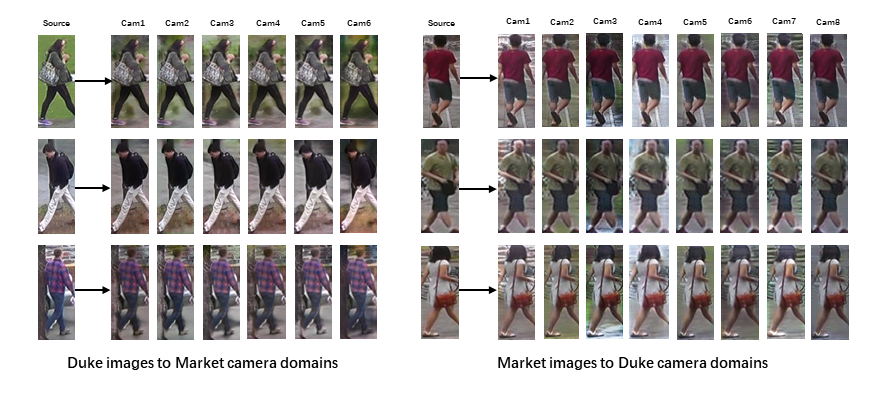}
\end{center}
\caption{Visual examples of images from source domain to target camera domains translation. The left seven columns map DukeMTMC-reID to Market-1501 six camera domains style. The right nine columns map Market-1501 to DukeMTMC-reID eight camera domains style. }
\label{fig:samples}
\end{figure*}

Recent works \cite{Liu2017Stepwise,fan2018unsupervised}  employ deep learning technology to estimate the labels of unlabeled target dataset. \cite{fan2018unsupervised} proposes an unsupervised approach which utilizes K-means to offer fake-labels for the unlabeled samples and train the re-ID model with those samples iteratively. \cite{Liu2017Stepwise} uses K-reciprocal nearest neighbors to get proximate label information for unsupervised video re-ID. A few unsupervised domain adaptation for person re-ID methods has proposed \cite{wang2018transferable,wei2017person,deng2018image,peng2016unsupervised,zhong2018generalizing}. Based on CycleGAN \cite{zhu2017unpaired},  SPGAN translates images from source domain to target domain via self-similarity \cite{deng2018image} which works by keeping the underlying identity information during translation. Then, the translated images are utilized to train re-ID model in a supervised way. Similar with \cite{deng2018image},  \cite{wei2017person} keep same contents during transferring. The above methods attempt to reduce the bias between source and target domain on image space and feature space, however they all ignore the divergence of image style caused by target camera domains. \cite{zhong2018camstyle} considers the intra-domain image variations caused by target cameras and learn discriminative representations of target domain. But this method cannot keep the same identity information between original images and translated images because identify semantic constraint is not considered. Furthermore, the above approaches have limited scalability in handling multiple domains since different models need to be trained on each pair of domains. Unlike them, our method explicitly considers the difference between source domain and target camera domains, and we can learn the relations among multiple domains using only a single model.

\section{Proposed Method} \label{Proposed Method}

\subsection{Baseline Re-ID Model}\label{sec:baseline}
The person re-ID task can be regarded as a classification problem. Thus, we use a classification model ResNet-50 \cite{he2016deep} as a backbone network for person re-ID. We use two FC layers to replace the final 1,000-dim fully connected(FC) layer after the Pooling-5 layer. The dimensions of the two FC layer are 1,024 and $N$, where $N$ is the number of classes in the dataset. The cross-entropy loss function is used to optimizing the model parameters. Our training process follows the ID discriminative embedding (IDE) \cite{zheng2016person}.

\begin{figure*}[t]
\setlength{\abovecaptionskip}{-0.2cm}
\setlength{\belowcaptionskip}{-0.3cm}
\begin{center}
\includegraphics[width=1.0\linewidth]{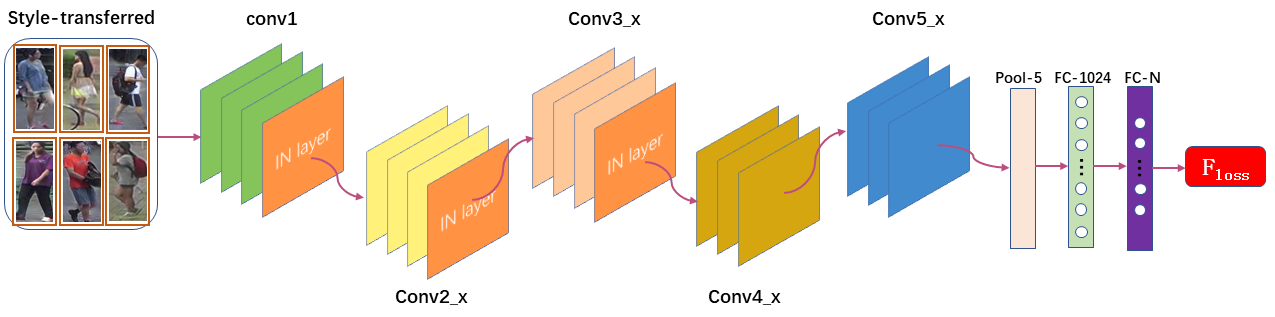}
\end{center}
\caption{We add three IN layers after the first convolution layer (conv1) and the first two convolution groups (conv2\_x, conv3\_x). Then we utilize the images which fit the style of target camera domains to train the IBN-reID model. The $F_{loss}$ is cross-entropy loss.}
\label{fig:IBN-framework}
\end{figure*}

\subsection{StarGAN Revisit}\label{StarGAN}
StarGAN \cite{choi2017stargan} is a novel and efficient generative adversarial network that learns the  mapping relations among multiple domains using only a single model. To make the generated images indistinguishable from real images, the adversarial loss is defined as:

\begin{equation}
\begin{split}
 \mathcal{L}_{adv}(x,c) = & \thinspace {\mathbb{E}}_{x} \left[ \log{{D}_\mathcal{T}(x)} \right]  \> \>  +   \\
 & \thinspace {\mathbb{E}}_{x, c}[\log{(1 - {D}_\mathcal{T}(G(x, c)))}],
\end{split}
\label{eq1}
\end{equation}
where $G$ generates an image $G(x, c)$ to fake $D$. The term ${D}_\mathcal{T}(x)$ is a probability distribution over sources. The goal of StarGAN is to translate x to an output image y which is classified as the target domain $c$. For this goal, a classifier is added on the top of $D$. The domain classification loss of real/fake images is defined as:
\begin{equation}
\mathcal{L}_{dom}^{r}(x,c') = {\mathbb{E}}_{x, c'}[-\log{{L}_{dom}(c'|x)}],
\label{eq2}
\end{equation}
\begin{equation}
\mathcal{L}_{dom}^{f}(x,c) ={\mathbb{E}}_{x, c}[-\log{{D}_{dom}(c|G(x, c))}].
\label{eq3}
\end{equation}
where the term ${D}_{dom}(c'|x)$ represents a probability distribution over domain labels computed by $D$. By minimizing Eq.\ref{eq2} and Eq.\ref{eq3}, $D$ learns to classify a real image $x$ to its original domain $c'$ and fake images to target domain $c$.
\par To guarantee that translated images preserve the content of their original images, StarGAN introduces a cycle-consistent loss \cite{zhu2017unpaired,kim2017learning} which is defined as:
\begin{equation}
\mathcal{L}_{rec}(x,c,c') = {\mathbb{E}}_{x, c, c'} [{||x - G(G(x, c), c')||}_{1} ],
\label{eq4}
\end{equation}

\subsection{IPGAN}\label{IPGAN}

One pedestrian has different appearances in different camera views and the bias is shown in Figure \ref{fig:carton}. In this paper, each camera domain in target domain is defined as a subdomain. In the real-world, it is entirely possible that the distribution of one subdomain may greatly differ from the distribution of the other ones, it is improper to take the target domain as a whole. A better way to smooth the bias between source and target domain is to learn image-image translation models that translate images from source domain to target camera domains rather than the whole target domain.
To this end, we propose the Identity Preserving Generative Adversarial Network (IPGAN). Our method can ensure the transferred image has a similar style as the style in target camera domain. The method is also able to keep the identity information of images from source domain during the translation. IPGAN consists of a style transfer model $G(x, c)$, a domain discriminator ${D}_{dom}$, and a semantic discriminator ${D}_{sem}$, as illustrated in Figure \ref{fig:IPGAN-framework}. The construction of IPGAN requires a source training set , the identity labels of source training set , a target training set, and the camera labels of target training set. Compared to identity labels, camera labels can be obtained along with surveillance videos without tedious and expensive manual labeling. There is no identity information of target set during image-image translation. Thus, IPGAN is an unsupervised learning method for person re-ID.

\textbf{Source to target camera domains Image-Image Translation.} In this work, we employ StarGAN \cite{choi2017stargan} to generate new training dataset without identity semantic constraint. Our goal is to train a single generator model that can translate images among the $L+1$ domains which consist of $L$ camera domains from target dataset and one source domain. The objective functions with respect to G and D are given, respectively, as

\begin{equation}
\mathcal{L}_{\mathcal{D}_{StarGAN}} =  - \mathcal {L}_{adv} +  {\lambda}_{dom}\thinspace\mathcal{L}_{dom}^{r},
\label{eq5}
\end{equation}
\begin{equation}
\mathcal{L}_{\mathcal{G}_{StarGAN}} =   \mathcal {L}_{adv} +  {\lambda}_{dom}\thinspace\mathcal{L}_{dom}^{f} +
{\lambda}_{rec}\thinspace\mathcal{L}_{rec},
\label{eq6}
\end{equation}

\par Specifically, following the training strategy in \cite{choi2017stargan}, the generator G contains 2 convolutional layers, 6 residual blocks and 2 transposed convolution layers. The discriminator $D_{dom}$ has the same structure as PatchGANs \cite{isola2017image}.

\setlength{\tabcolsep}{10pt}
\begin{table*}[t]
\begin{center}
\begin{tabular}{l|cccc|cccc}
\hline
\multicolumn{1}{c|}{\multirow{2}{*}{Methods}}&\multicolumn{4}{c|}{Market-1501}&\multicolumn{4}{c}{DukeMTMC-reID}\\
\cline{2-9}
\multicolumn{1}{c|}{}&rank-1&rank-5&rank-10&mAP&rank-1&rank-5&rank-10&mAP\\
\hline
Supervised Learning &84.8&93.7 &96.2 &65.3&72.6&84.5&87.6&52.0\\
Baseline+Direct Transfer &44.3&61.9&69.7&18.4 &30.2&45.1&51.5 &16.1\\
\hline
Baseline+CycleGAN \cite{deng2018image} &49.9&67.1 &74.2 &22.6&39.2&54.8&60.5&20.1\\
\hline
Baseline+CamStyle \cite{zhong2018camstyle} &58.8&78.2 &84.3 &27.4&48.4&62.5&68.9&25.1\\
Baseline+StarGAN &\textbf{54.3}&\textbf{74.7}&\textbf{81.5}&\textbf{24.4}&\textbf{44.1}&\textbf{60.0}&\textbf{65.8}&\textbf{21.9}\\
{Baseline+IPGAN}&\textbf{56.4}&\textbf{76.0}&\textbf{82.5}&\textbf{25.6}&\textbf{46.8}&\textbf{62.0}&\textbf{67.9}&\textbf{26.7} \\
\hline
\end{tabular}
\end{center}
\setlength{\abovecaptionskip}{0cm}
\setlength{\belowcaptionskip}{-0.1cm}
\caption{ We use baseline for feature learning. Methods comparison using DukeMTMC-reID/Market-1501  as source, and Market-1501/DukeMTMC-reID as target. ¡°Supervised learning¡± denotes training and testing on target domain, simultaneously. ¡°Direct Transfer¡± means directly applying the model trained by useing images from source domain on the target domain. CycleGAN is used to translate images from source domain to target domain. CamStyle, StarGAN, and IPGAN are used to translate images from source domain to target camera domains. } \label{Compare base}
\label{table:1}
\end{table*}

\textbf{Identity semantic constraint loss function.} As analyzed in the above, we aim to preserve the identity information of images from source domain during image-image translation. To fulfill this goal, we design a semantic discriminator $D_{sem}$, which is used for identity preserving. The consistency on person identity is important for person re-ID training. The structure of $D_{sem}$ is similar to the baseline(Section \ref{sec:baseline}). The identity semantic constraint loss function is:
\begin{equation}
\mathcal{L}_{sem}(x_{s},y) ={\mathbb{E}}_{x_{s}, y}[-\log{{D}_{sem}(y|G(x_{s}, c))}],
\label{eq7}
\end{equation}where $G$ generates an image $G(x_{s}, c)$ conditioned on both the input source domain images $x_{s}$ and the target camera domain label $c$. A $x_{s}$ corresponds to a identity label $y$. In a min-batch, ${D}_{sem}$ aims to classify a generated $G(x_{s}, c)$ to its corresponding original identity label $y$.

\par The above two parts are integrated in a framework IPGAN. Its structure is $\big\{G, ({D}_{dom} ,{D}_{sem})\big\}$ as illustrated in Figure \ref{fig:IPGAN-framework}. The generator $G$ maps source domain images to the styles of target camera domains. The discriminator ${D}_{dom}$ is used to distinguish real and fake images and judge the domain that the translated images belong to. The ${D}_{sem}$ enforces the images from source domain keeping identity information after translation. Note that ${D}_{sem}$ is a pretrained classifier with source domain training set and only used to optimize $G$. Only the identity information from source domain is applied. When we train the IPGAN, the parameters of ${D}_{sem}$ is fixed.
\par Finally, the overall IPGAN objective function can be written as:
\begin{equation}
\mathcal{L}_{\mathcal{D}_{IPGAN}} =  - \mathcal {L}_{adv} +  {\lambda}_{dom}\thinspace\mathcal{L}_{dom}^{r},
\label{eq8}
\end{equation}
\begin{equation}
\begin{split}
\mathcal{L}_{\mathcal{G}_{IPGAN}} = & \thinspace \mathcal {L}_{adv} + {\lambda}_{dom}\thinspace\mathcal{L}_{dom}^{f}\> \>  +  \\
 & \thinspace {\lambda}_{rec}\thinspace\mathcal{L}_{rec} + {\lambda}_{sem}\thinspace\mathcal{L}_{sem},
\end{split}
\label{eq9}
\end{equation}
where  ${\lambda}_{dom}$, ${\lambda}_{rec}$, and ${\lambda}_{sem}$ are three hyper-parameters, that tradeoff the importance of domain classification, reconstruction loss, and identity semantic loss.
We use ${\lambda}_{dom}$ = 1, ${\lambda}_{rec}$ =10, and ${\lambda}_{sem}$ = 1 in our experiments.

\par With the learned IPGAN model, for an image in source domain we generate $L+1$ fake samples whose styles are similar to $L+1$ domains, and meanwhile the identity information is keeping during the Image-Image translation. Note that the $L+1$ samples contain a fake image whose style is same as its original style and the fake image should be ignored. Finally, we train re-ID model with the style transferred images in a supervised way.

\subsection{IBN-re-ID model}\label{IBN-re-ID model}

\par To address the problem of the appearance gap between source and target camera domains, a intuitive way is to introduce appearance invariance into re-ID models. Inspired by \cite{pan2018two}, we propose a novel deep network named IBN-reID for person re-ID. In IBN-reID, the appearance variance is mainly reflected in shallow features, and the change of content information is reflected in deep features. The characteristic of IBN-reID is that batch normalization and instance normalization are utilized together in the framework. The advantage of batch normalization is preserving discrimination between individual samples by deep features, but the drawback is that it makes CNNs vulnerable to appearance transforms. On the contrary, instance normalization eliminates individual contrast, but diminishes useful information at the same time. Instance normalization and batch normalization are integrated as IBN-block which learns to capture and eliminate appearance variance, while maintains discrimination of the learned features. To our best knowledge, it's the first attempt to introduce IBN-block into person re-ID. IBN-reID has better generalization ability than baseline models, in the case of deploying the model trained on source domain to target domain without any domain adaptation.
\par We use ResNet-50 \cite{he2016deep} as base model that consists of four groups of residual blocks. We add three IN layers after the first convolution layer (conv1) and the first two convolution groups (conv2\_x, conv3\_x), respectively. The latter layers are the same as baseline model (Section \ref{sec:baseline}).

\section{Experiment}

\subsection{Datasets}

Two widely used benchmark datasets are chosen for experiments, Market-1501 \cite{zheng2015scalable} and DukeMTMC-reID \cite{zheng2017unlabeled,ristani2016performance} because both datasets 1) are large-scale and 2)  camera labels for each image is available.
\par \textbf{The Market-1501}\cite{zheng2015scalable} dataset contains 32,668 images from 1501 identities collected from 6 cameras. All of the images are produced by deformable part mode (DPM) \cite{felzenszwalb2010object}. The dataset is split into two non-over-lapping parts: 12,936 images from 751 identities for training and 19,732 images from 750 for testing. Moreover, 3,368 query images from 750 identities are used to retrieve the matching persons in the gallery.
\par \textbf{The DukeMTMC-reID} \cite{zheng2017unlabeled,ristani2016performance} is also a large-scale re-ID dataset which is collected from 8 cameras. It contains
16,522 training images from 702 identities, 2,228 query images from another 702 identities and 17,661 gallery images, 36411 images belonging to 1404 identities in total.
\par During the course of the experiment, training sets of Market-1501 and DukeMTMC-reID are used for style transfer. We use rank-1 accuracy and mean average precision (mAP) for evaluation on both datasets.

\subsection{Implementation Details}
 We use Pytorch to train IPGAN on NVIDIA GeForce GTX Titan xp GPU using the training set of Market-1501 and DukeMTMC-re-ID. We  use a single model to learn the mapping between source domain and target camera domains.  For the generator network, we use instance normalization in all layers except the last output layer. For the discriminator $D_{dom}$ network, we use Leaky ReLU with a negative slope of 0.01. The structure of discriminator $D_{sem}$ is simailr to baseline model. The input images are resized to $128 \times 64$. The learning rate is 0.0001 at the first 100 epochs and linearly reduces to zero for the last 100 epochs. for each image in the source training set, we generate $L$ style-transferred images (the number of target cameras). These fake images are regarded as a new training set which is used to train re-ID model.

\setlength{\tabcolsep}{10pt}
\begin{table*}[t]
\begin{center}
\begin{tabular}{l|cc|cc}
\hline
\multicolumn{1}{c|}{\multirow{2}{*}{Methods}}&\multicolumn{2}{c|}{Market-1501}&\multicolumn{2}{c}{DukeMTMC-reID}\\
\cline{2-5}
\multicolumn{1}{c|}{}&Real Image Accuracy&Fake Image Accuracy&Real Image Accuracy&Fake Image Accuracy\\
\hline
\hline
StarGAN &1.0&0.22 &1.0&0.28\\
\hline
{IPGAN}&1.0&\textbf{0.99}&1.0&\textbf{0.97}\\
\hline
\end{tabular}
\end{center}
\setlength{\abovecaptionskip}{0cm}
\setlength{\belowcaptionskip}{-0.1cm}
\caption{The classification accuracy of new Market-1501/DukeMTMC-reID dataset generated by StarGAN is 0.22/0.28. The classification accuracy of new Market-1501/DukeMTMC-reID dataset generated by IPGAN is  0.99/0.97.} \label{Compare base}
\label{table:2}
\end{table*}

\setlength{\tabcolsep}{10pt}
\begin{table*}[t]
\begin{center}
\begin{tabular}{l|cccc|cccc}
\hline
\multicolumn{1}{c|}{\multirow{2}{*}{Methods}}&\multicolumn{4}{c|}{Market-1501}&\multicolumn{4}{c}{DukeMTMC-reID}\\
\cline{2-9}
\multicolumn{1}{c|}{}&rank-1&rank-5&rank-10&mAP&rank-1&rank-5&rank-10&mAP\\
\hline
\hline
Baseline+Direct Transfer &44.3&61.9 &69.7 & 18.4&30.2&45.1&51.5&16.1\\
Baseline+StarGAN &54.3&74.7 &81.5 & 24.4&44.1&60.0&65.8&21.9\\
Baseline+IPGAN &56.4&75.6 &82.5 & 25.6&46.8&62.0&67.9&25.7\\
\hline
IBN-reID+Direct Transfer &\textbf{45.7}&\textbf{62.8}&\textbf{70.8}&\textbf{19.8}&\textbf{32.4}&\textbf{47.4}&\textbf{54.6}&\textbf{17.3}\\
IBN-reID+StarGAN &\textbf{56.0}&{74.6}&\textbf{81.5}&\textbf{25.5}&\textbf{44.6}&\textbf{60.0}&\textbf{66.0}&\textbf{22.3}\\
{IBN-reID+IPGAN} &\textbf{57.2}&\textbf{76.0}&\textbf{82.7}&\textbf{28.0}&\textbf{47.0}&\textbf{63.0}&\textbf{68.1}&\textbf{27.0}\\
\hline
\end{tabular}
\end{center}
\setlength{\abovecaptionskip}{0cm}
\setlength{\belowcaptionskip}{-0.1cm}
\caption{Comparison of various domain adaptation methods over Baseline model and IBN-reID model. The best results are in \textbf{bold}} \label{Compare base}
\label{table:3}
\end{table*}

 \subsection{Evaluation}
 \textbf{Baseline accuracy.} When the re-ID model is baseline model. The supervised learning method and the direct transfer method are specified in Table \ref{table:1}. When trained and tested both on the target set, excellent accuracy can be achieved. However, when trained on source dataset and directly tested on target dataset, the performance drops significantly. For instance, the baseline model trained and tested on Market-1501 achieves $84.8\%$ in item of rank-1 accuracy, but drops to $44.3\%$ when trained on DukeMTMC-reID training set and tested on Market-1501 testing set. The reason is the \emph{bias} of data distributions between different domains.

\setlength{\tabcolsep}{11pt}
\begin{table*}[t]
\begin{center}
\begin{tabular}{l|cccc|cccc}
\hline
\multicolumn{1}{c|}{\multirow{2}{*}{Methods}}&\multicolumn{4}{c|}{Market-1501}&\multicolumn{4}{c}{DukeMTMC-reID}\\
\cline{2-9}
\multicolumn{1}{c|}{}&rank-1&rank-5&rank-10&mAP&rank-1&rank-5&rank-10&mAP\\
\hline
\hline
LOMO \cite{liao2015person} &27.2&41.6 &49.1 & 8.0&12.3&21.3&26.6&4.8\\
BOW \cite{zheng2015scalable} &35.8&52.6 &60.3 & 14.8&17.1&28.8&34.9&8.3\\
\hline
UMDL \cite{peng2016unsupervised} &34.5&52.6 &59.6 & 12.4&18.5&31.4&37.6&7.3\\
PUL \cite{fan2018unsupervised} &45.5&60.7 &66.7 & 20.5&30.0&43.4&48.5&16.4\\
CAMEL \cite{yu2017cross} &54.5&- &- & 26.3&-&-&-&-\\
\hline
PTGAN \cite{wei2017person} &38.6&- &66.1 & -&27.4&-&50.7&-\\
SPGAN \cite{deng2018image} &51.5&70.1 &76.8 & 22.8&41.1&56.6&63.0&22.3\\
TJ-AIDL \cite{wang2018transferable} &58.2&74.8 &81.1 & 26.5&44.3&59.6&65.0&23.0\\
CamStyle \cite{zhong2018camstyle} &\textbf{58.8}&\textbf{78.2} &\textbf{84.3} & 27.4&\textbf{48.4}&62.5&\textbf{68.9}&25.1\\
\hline
IPGAN &56.4&75.6 &82.5 & 25.6&46.8&62.0&67.9&25.7\\
{IPGAN+IBN-reID} &57.2&76.0&82.7&\textbf{28.0}&47.0&\textbf{62.8}&68.1&\textbf{27.0}\\
\hline
\end{tabular}
\end{center}
\setlength{\abovecaptionskip}{0cm}
\setlength{\belowcaptionskip}{-0.1cm}
\caption{Comparison with state-of-the-art on Market-1501 and DukeMTMC-reID. The best results are in \textbf{bold}} \label{Compare base}
\label{table:4}
\end{table*}

\par \textbf{The impact of ¡°source domain to target camera domains Image-Image Translation¡±.} Firstly, following the work of \cite{deng2018image}, we use CycleGAN \cite{zhu2017unpaired} to translate the labeled images from the source domain to the target domain and then train the baseline re-ID model with the translated images in a supervised way. As show in Table \ref{table:1}. When trained on DukeMTMC-reID training set and tested on Market-1501 testing set, rank-1 accuracy improves from $44.3\%$ to $49.9\%$ and mAP accuracy improves from 18.4 to 22.6. when trained on Market-1501 training set and tested on DukeMTMC-reID testing set, rank-1 accuracy improves from $30.2\%$ to $39.2\%$ and mAP accuracy improves from $16.1$ to $20.1$.

\par
\begin{figure}[t]
\setlength{\abovecaptionskip}{-0.2cm}
\setlength{\belowcaptionskip}{-0.3cm}
\begin{center}
\includegraphics[width=0.9 \linewidth]{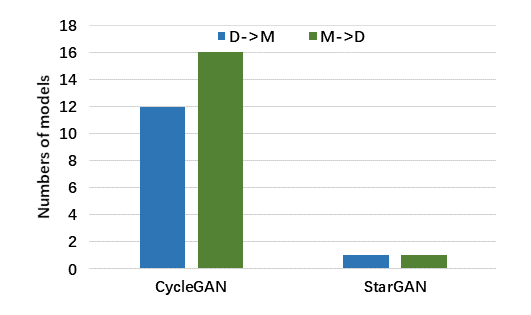}
\end{center}
\caption{Style transfer from source domain to target camera domains with CyleGAN and IPGAN}
\label{fig:fig5}
\end{figure}

 Secondly, we consider the bias between source domain and target camera domains. Very resently, CamStyle \cite{zhong2018camstyle} is a similar work with ours and implement the translation from source domain to target camera domains with CycleGAN \cite{zhu2017unpaired}, while we use StarGAN \cite{choi2017stargan} to implement it. As show in Table \ref{table:1}, compared to CamStyle\cite{zhong2018camstyle}, despite the results of ``Baseline+StarGAN'' are inferior to it, ``Baseline+StarGAN'' is more efficient. As show in Figure \ref{fig:fig5}, when translate from Market-1501(source domain) to DukeMTMC-reID(8 camera domains), CamStyle \cite{zhong2018camstyle} needs 16 pairs of \big\{G, D\big\} (CycleGAN has two generator-discriminator pairs). When translation from DukeMTMC-reID(source domain) to Market-1501(6 camera domains), it needs 12 pairs of \big\{G, D\big\} on Market-1501. StarGAN \cite{choi2017stargan} only uses two pairs of $\big\{G, {D}_{dom}\big\}$ to complete it. Results are showed in Table \ref{table:1}.  Compared to the ``Baseline+CycleGAN'', the ``Baseline+StarGAN'' gains +$4.4\%$ improvements in item rank-1 accuracy and +$1.8$ in item of mAP when tested on Market-1501. When tested on DukeMTMC-reID, the performance gains +$4.9\%$ in item of rank-1  accuracy and +$1.8$ in item of mAP. Through such source to target camera domains adaptation method, effective improvement can be achieved.

 \par \textbf{The impact of IPGAN.} Compared to StarGAN \cite{choi2017stargan} , IPGAN is same efficiency as StarGAN \cite{choi2017stargan} but owns identity semantic constraint. We conduct experiment to verify the influence of the identity semantic constraint. As show in Table \ref{table:2}, we analyze the difference of images translated by StarGAN and IPGAN. We use DukeMTMC-reID/Market-1501 training set to train two classifiers which are used to predict the identities of translated images. The classification accuracy of DukeMTMC-reID/Market-1501 training set after translated by StarGAN is only $28.0\%/22.0\%$. But, when DukeMTMC-reID/Market-1501 training set is translated by IPGAN, the classification accuracy is $97.0\%/99.0\%$. The fake images generated by IPGAN keep almost the same identity information as the original real images. However, most of the images generated by StarGAN lose identity semantic information. The performance of re-ID is showed in Table \ref{table:1}. ``Baseline+IPGAN'' gains a better performance than ``Baseline+StarGAN'' in item of rank-k and mAP accuracy.

\par \textbf{The impact of ¡°IBN-block¡±.} To further improve re-ID performance on target dataset, we propose IBN-re-ID model (Figure \ref{fig:IBN-framework}). As show in Table \ref{table:3}, compared to methods with baseline re-ID model, when we use IBN-reID model as re-ID model,  the rank-k and mAP accuracy are improved in varying degree. More obviously ``IBN-reID+Direct Transfer'' gains rank-1 accuracy in $32.4\%$ and mAP in 17.3 when tested on DukeMTMC-reID, surpassing the ``Baseline+Direct Transfer'' by +$2.2\%$ and +$1.2$, respectively. The similar improvement is obtained when tested on Market-1501. However, compared with ``Baseline+StaGAN/IPGAN'', the models with IBN-reID, i.e., ``IBN-reID+StarGAN/IPGAN'', get slight improvements in term of rank-1 accuracy. The main reason is that the \emph{bias} between source and target camera domains is significantly reduced by StarGAN and IPGAN. Thus, in IBN-reID, the instance normalization only provide limited helps which eliminate appearance variance in shallow layers. This weaken the generalization capacity of IBN-reID. Even so, the mAP values have been significantly improved and the models get the best results.

\par \textbf{Comparison with State-of-the-art Methods.} We compare the proposed method with the state-of-the-art unsupervised learning methods. Table \ref{table:4} presents the comparison when tested on Market-1501 and DukeMTMC-reID. Firstly, we compare with two hand-crafted features: Bag-of-Words(BoW) \cite{zheng2015scalable} and local maximal occurrence (LOMO)\cite{liao2015person}. Their inferiority is obvious. Secondly, we compare the proposed methods with three unsupervised methods, including CAMEL \cite{yu2017cross}, PUL \cite{fan2018unsupervised}, and UMDL \cite{peng2016unsupervised}. Finally we also compare four unsupervised domain adaptation approaches, including PTGAN \cite{wei2017person}, SPGAN(+LMP) \cite{deng2018image}, TJ-AIDL \cite{wang2018transferable} and CamStyle \cite{zhong2018camstyle}. Comparing with those methods, when tested on Market-1501, The proposed method achieves rank-1 accuracy = $57.2\%$ and \textbf{the best mAP = 28.0}. When tested on DukeMTMC-reID, The result achieves rank-1 accuracy = $47.0\%$ and \textbf{the best mAP = 27.0}. Compared with CamStyle \cite{zhong2018camstyle}, the proposed method achieves a similar rank-k accuracy and slight improvement in item of mAP. Especially when tested on DukeMTMC-reID, we gain $+1.9$ higher in mAP. The proposed method achieves a very competitive re-ID accuracy on ``Market-1501 to Duke'' and ``DukeMTMC-reID to Market-1501''.

\section{Conclusion}
In this paper, we propose the identity preserving generative adversarial network (IPGAN), a novel and efficient unsupervised domain adaptation method for person re-identification. IPGAN is used to translate images from source domain to target camera domains with identities preserving and thus we have a new training set. Then, we use the new training set to train re-ID model. Moreover, to obtain better performance, we propose IBN-reID model, which has better generalization ability than baseline model. Experiments on the Market-1501 and DukeMTMC-reID datasets show that we can achieve more competitive performance. In the feature, we will propose a new method to solve identity semantic constraint.

{\small
\bibliographystyle{ieee}
\bibliography{egbib}
}

\end{document}